\pdfoutput=1
\documentclass[sigconf]{acmart}

\usepackage{booktabs}
\usepackage{amsmath}
\usepackage{bbm}
\AtBeginDocument{%
  \providecommand\BibTeX{{%
    \normalfont B\kern-0.5em{\scshape i\kern-0.25em b}\kern-0.8em\TeX}}}

\usepackage{todonotes}

\begin{document}

\title{Explainable Health Risk Predictor with Transformer-based Medicare Claim Encoder}

\author{Chuhong Lahlou}
\authornote{Contact author}
\email{lahlou\_chuhong@bah.com}
\affiliation{%
  \country{Booz Allen Hamilton}
}

\author{Ancil Crayton}
\email{crayton\_ancil@bah.com}
\affiliation{%
  \country{Booz Allen Hamilton}
}

\author{Caroline Trier}
\email{trier\_caroline@bah.com}
\affiliation{%
  \country{Booz Allen Hamilton}
}
\author{Evan Willett}
\email{willett\_evan@bah.com}
\affiliation{%
  \country{Booz Allen Hamilton}
}


\begin{abstract}
  In 2019, The Centers for Medicare and Medicaid Services (CMS) launched an Artificial Intelligence (AI) Health Outcomes Challenge\footnote{\href{https://ai.cms.gov/}{\textit{ai.cms.gov}}} seeking solutions to predict risk in value-based care for incorporation into CMS Innovation Center payment and service delivery models. Recently, modern language models have played key roles in a number of health related tasks. This paper presents, to the best of our knowledge, the first application of these models to patient readmission prediction. To facilitate this, we create a dataset of 1.2 million medical history samples derived from the Limited Dataset (LDS) issued by CMS. Moreover, we propose a comprehensive modeling solution centered on a deep learning framework for this data. To demonstrate the framework, we train an attention-based Transformer to learn Medicare semantics in support of performing downstream prediction tasks thereby achieving 0.91 AUC and 0.91 recall on readmission classification. We also introduce a novel data pre-processing pipeline and discuss pertinent deployment considerations surrounding model explainability and bias. 
\end{abstract}





\maketitle

\section{Introduction}

Value-based care increasingly requires healthcare providers to manage both quality and total cost for defined patient groups. To optimize for success, providers develop methods to understand the future health risks of their patients in order to allocate resources in ways that deliver efficient care. While there are multiple definitions of health risk, we focus in this paper on the risk of unplanned readmission which we define as \textit{the probability of a patient  being unexpectedly readmitted to an inpatient facility within 30 days of discharge}. The Centers for Medicare \& Medicaid Services (CMS) tracks and publicly reports this metric as a measure of a hospital’s care standards and factors into that hospital's allowable reimbursements \cite{Roy2015}. Health systems can provide preventive services and engage patients in educational programs to prevent unplanned readmissions. This strategy can be prohibitively costly to both individuals and health systems. Therefore, it is preferable to focus resources on only those patients most at risk.

CMS recently reported the estimated cost of unplanned readmissions is roughly \$17.9 billion per year \cite{Roy2015}. Hence, tackling this problem is not only important for addressing future health risks of individuals but also for lowering the total cost of unplanned readmissions. To curtail this problem, the Patient Protection and Affordable Care Act introduced penalties to hospitals with excessive readmission rates at a minimum of 3\% of their Medicare reimbursement. However, these measures have not yet been deemed sufficient  to reverse the problem \cite{Xiao2018}.

An alternative approach is to estimate and inform healthcare providers the health risk associated with each patient at discharge. This enables healthcare providers to determine and take commensurate preventative actions for patients who would most likely benefit from targeted interventions. This paper assists in this approach by leveraging deep learning and predictive analytics to estimate each patients health risk with actionable explainability to health care professionals. With no pre-existing public datasets available for such a task, we created a custom dataset of 1.2 million medical history and readmission outcomes based on Medicare claims records made available by CMS through the AI Health Outcomes Challenge.  

In this paper, we first review in Section~\ref{sec:related-work} recent advances in deep learning health care applications. In Section \ref{sec:approach}, we present the technical approach, including a data pre-processing pipeline for medical claims, a deep learning framework, and a means for providing explainability. Next, in Section~\ref{sec: eval}, we demonstrate the framework by training an example model to predict unplanned hospital readmissions within 30 days of discharge for Medicare beneficiaries. Finally, in Section~\ref{sec:op}, we discuss a number of considerations 
surrounding production deployment including operationalization,  generalizability, and bias mitigation.

\section{Related Work}\label{sec:related-work}
Our paper relates broadly to the literature on using deep learning models for health applications with unstructured data. As research in this area is vast and continuously growing, we direct readers to learn about this literature through the  in-depth surveys of deep learning approaches for various downstream applications using electronic health records (EHRs) by \cite{Shickel2018} and \cite{Si2020}.

A more narrow focus, this paper relates to using modern language models, specifically Transformer-based models, to study health-related tasks. A significant portion of these studies focus on modeling EHRs for learning representations and performing downstream tasks. There have been applications of language models to learn both from structured data and general clinical language representations \cite{Alsentzer2019,Steinberg2021,Si2019,KS2019}. BEHRT is a BERT model trained on EHRs to learn general representations that scale well across a wide range of downstream tasks \cite{Li2020}. Med-BERT is a BERT model trained on EHRs for disease prediction \cite{Rasmy2020}. G-BERT is a graph-augmented Transformer model used to develop medication recommendations \cite{Shang2019}. Although large-scale language models have been used to study various health-related tasks, we present, to the best of our knowledge, the first application of a Transformer-based model to predict patient readmission using medical claims data.

More specific to our application, our paper relates to the literature developing empirical methods for predicting unplanned hospital readmissions. \cite{Artetxe2018} survey prediction methods, covering traditional statistical models (e.g. regression) and machine learning models. In this study, it is reported that the AUC prediction performance in the literature ranges from 0.54-0.92. The highest performing approach, achieving an AUC of 0.92, leverages support vector machines and LASSO regression to predict 30-day patient readmission due to heart failures. We distinguish our results from this study by achieving a 0.91 AUC on \textit{general} patient readmission (not dependent on a diagnosis) while also allowing for individual explainability through our model choice.

Our paper also relates to the literature on using deep learning to predict patient readmission. Past research has developed various types of deep learning models to predict patient readmission across various forecast horizons. Most approaches follow modeling the sequential structure in EHRs by using a recurrent neural network (RNN). \cite{Rajkomar2018} predict 30-day unplanned readmission using a large set of EHRs achieving AUC of around 0.75-0.76, and \cite{Choi2016a} modeled the structured information in EHRs such as medical codes and frequency of visits by using an RNN approach.Other studies have followed the approach of using embedding models for learning representations of longitudinal EHR data and leveraging these vector representations to inform readmission predictions. \cite{Xiao2018} used deep contextual embeddings of clinical concepts learned from EHRs to predict patient readmission using a Topic Recurrent Neural Network (TopicRNN). \cite{Zhang2018} trained an embedding model, Patient2Vec, on EHRs that allows for population- and individual-level interpretability and achieved an AUC of 0.79 for predicting patient readmission within a 6-month period. Taking a different approach to RNNs and embedding models, \cite{Yang2017} developed TaGiTeD, a tensor decomposition method trained to predict 1-year hospital readmissions and allow for interpretable phenotype representation while achieving a classification performance of between 0.68-0.78 AUC.

As far as we are aware, our work is the first use of deep contextual embeddings of medical claims to predict patient readmission using a Transformer-based approach. We find this advantageous as we would like to model the sequential structure of our data. Additionally, the Transformer model was primarily designed to overcome the limitations of recurrent neural networks (e.g. exploding/vanishing gradients for long sequences, sequential training, etc.) \cite{Vaswani2017}, which a significant portion of the literature relies on, while being able to leverage the attention mechanisms to provide individual-level interpretability of predictions while previous approaches focus more on global interpretability.

Finally, our study relates to the literature on patient readmission leveraging survival analysis. Survival analysis is performed when there is an interest in the time to an event. Relevant to our work, this would be analyzing the time to an unplanned hospital readmission following the discharge of a patient. The literature in this area relies on using random survival forests \cite{Ishwaran2008} and Cox regressions \cite{Cox1972}. This literature covers various horizons and types of unplanned patient readmission (ranging from general unplanned patient readmission to diagnosis-specific readmissions). The random survival forest approaches have an AUC performance range of 0.69-0.72 \cite{Hao2015, Padhukasahasram2015}, while the Cox regression approaches show an AUC range of 0.65-0.82 \cite{Corrigan1992, Jencks2009, Wang2012, A.2015, V.2015, Padhukasahasram2015, Pereira2015, Tulloch2016, Yu2015, Krumholz2016}. While survival analysis is an appropriate model for our data, the context of our task is focused on a single survival point: the 30-day window. For this reason we prefer to model the single split as a classification task as it falls out naturally, and allows the model to focus all representational capacity toward this one financially and legally important decision point. A survival approach would involve modeling the entire distribution of readmissions, which entails calibrating concerns that are non-trivial to resolve, and does not provide any additional benefit for the purpose of our application.


\section{Technical Approach} \label{sec:approach}

Our approach consists of three primary components: data pre-processing, a Transformer model, and AI explainability. Specifically, we draw upon natural language processing techniques to identify key patterns in claims history, stratify beneficiaries, and identify potential interventions at the beneficiary level. This approach enables new features to be learned from heterogeneous claims variables (including categorical and numerical fields) before sending them to the downstream classification model to make risk predictions. In this section, we discuss each component in detail.

\subsection{Data Pre-processing}
Medicare claims data consist of multiple records per patient with varying lengths of non-human readable diagnostic and procedural codes. To normalize this data, we built an ingestion and transformation pipeline. The CMS Limited Dataset (LDS) that was used for the AI Health Outcomes Challenge consists of a 5\% sub-sample of all Medicare claims from 2008-2011.  The LDS dataset contains seven different claim types, including Durable Medical Equipment (DME), Carrier (CAR), Home Health Aide (HHA), Hospice (HOSP), Inpatient (INP), Outpatient (OUT), Skilled Nursing Facility (SNF), and a denominator (DENOM) file. Each of these claim types consist of the data elements necessary to describe a beneficiary's claim for services through Medicare.  Different claim types have different data schemas and data elements.  For example, an Outpatient claim describes the facility where the service took place (whether it was a hospital, a physician's office, or another type of facility), the type of services that were provided, the dates of the provided services, any applicable diagnosis or procedure codes, and the billing codes associated with the provided services. On the other hand, a Carrier claim describes the services provided by the physician or healthcare provider (who may work at an inpatient or outpatient facility), including identifiers for the supplier of the services, the procedures provided by the supplier, and the billing codes associated with the physician-provided services.  Additionally, the CMS system update in 2011 resulted in a schema change in selected claim files\footnote{\textit{https://www.cms.gov/Research-Statistics-Data-and-Systems/Files-for-Order/LimitedDataSets/StandardAnalyticalFiles} has a complete description of the contents of the CMS Limited Dataset}.

\begin{figure*}
    \centering
    \includegraphics[width=\textwidth, keepaspectratio]{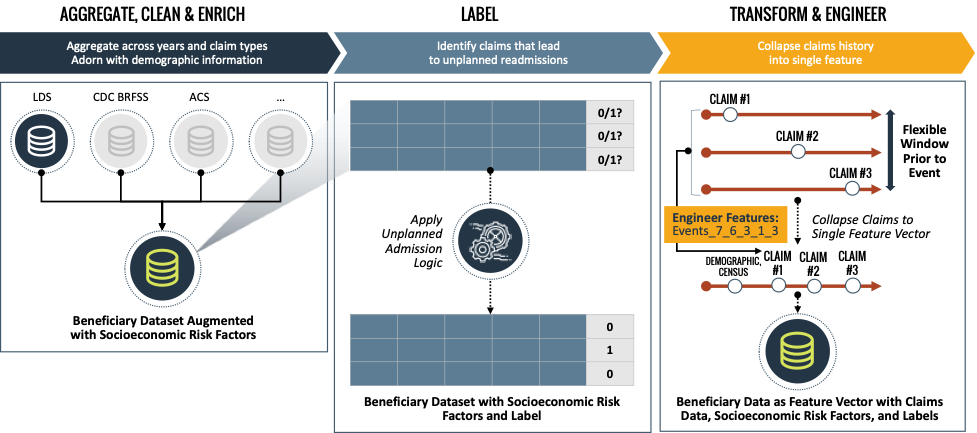}
    \caption{The data pre-processing pipeline used to normalize Medicare and Census data: (1) data aggregation, (2) label generation, and (3) data transformation.}
    \label{fig:dp}
\end{figure*}

As illustrated in Figure~\ref{fig:dp}, the pipeline loads the Medicare claims data from the Limited Dataset (LDS) and augments it with publicly available Census data keyed off facility county codes, annotates unplanned readmission event labels, builds patient event history, builds a dictionary, and finally performs data transformations and tokenizations.

\subsubsection{Data Aggregation}

In order to have a single system of record for the downstream use of data, we leveraged an SQL database for data staging and schema consolidation.

\begin{figure}
    \centering
    \includegraphics[width=\columnwidth, keepaspectratio]{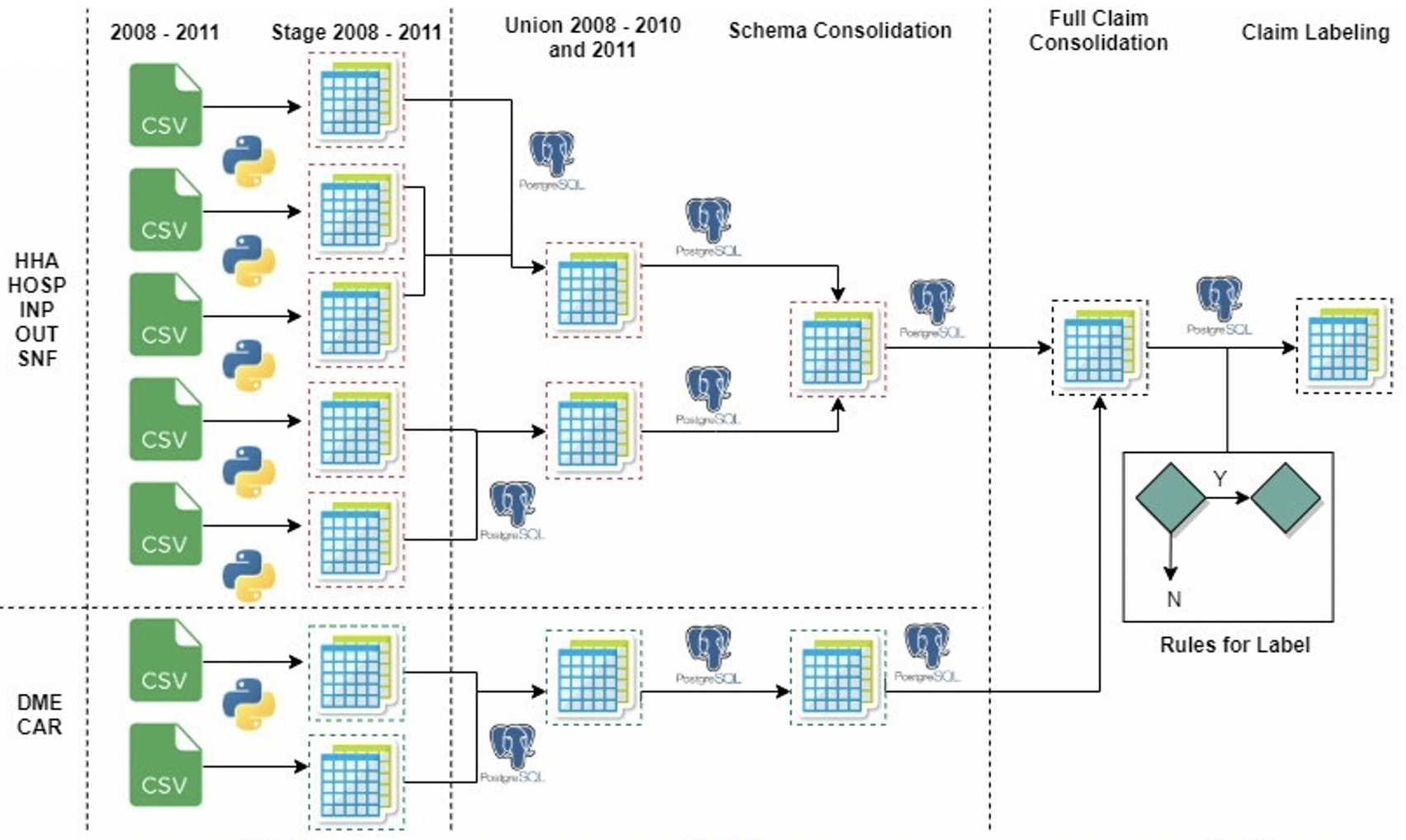}
    \caption{The processing flow diagram for the data aggregation and labeling steps: (1) staging by type, (2) schema consolidation, and (3) label assignment.}
    \label{fig:lds_load}
\end{figure}

As illustrated in Fig.\ref{fig:lds_load}, LDS claim files are loaded and staged in separate tables by type in Persistent Staging Areas. Then, for each claim type, all claims are converted into the 2011 schema and inserted into the database with expanded and filtered forms. The filtered representations were down-sampled to contain key attributes selected by clinical experts\footnote{For details, please contact the authors.}. County-level Census and other demographic information was used to enrich the LDS claims using county and state IDs. This information is sourced from Behavioral Risk Factor Surveillance System (BRFSS)\footnote{See \href{https://www.cdc.gov/brfss/}{\textit{cdc.gov/brfss}}} and the American Community Survey (ACS).

\subsubsection{Data Labeling}

We label the claims data according to the following rules: if the difference between an INP claim discharge date and the next INP claim admission date is less or equal to 30 days then the claim is labeled ``1'' for \textit{leading to readmission}. Note that if an INP claim's discharge status indicates a transfer of care, i.e. same-day discharge and admission, then the label is also ``1''. In all other cases, the claims are labeled ``0'' indicating \textit{not leading to readmission}. The final dataset contains 334M total claims of which 10.9M were INP claims, and 2.4M had label ``1''.

\subsubsection{Data Transformation}

To build data samples representing patient event history for model training, the filtered samples containing the clinically relevant variables are extracted from the database, converted into a standard format, and then exported into text sequences. 

As depicted by Step 3 in Fig.\ref{fig:dp}, a data sample contains multiple segments with variables from distinct sources and medical claims. The sample starts with a patient's personal information which includes date of birth, gender, reason for entitlement, etc.. The next set of variables is the county-level demographic data associated with the INP claim facility. Following that, clinically relevant variables from the patient's last 3 months of claims are sequenced from the least to most recent, with time ranges indicating the duration between each claim relative to the current INP claim. The final set of variables are the clinically relevant variables from the current INP claim for which the readmission label is assigned.

While the encoding scheme we use is able to handle free text, categorical, and numerical data, a predefined vocabulary is required for vectorization. Due to the nature of granular medical codes, and even with relatively coarse quantization of floating point values, the vocabulary size grows significantly with additional claims data making the semantic embedding learning computationally challenging. In order to limit the vocabulary size, we performed transformations on the dataset that includes variable name grouping, ICD\footnote{https://www.cdc.gov/nchs/icd/icd10cm.htm} and HCPCS/\footnote{https://www.cms.gov/Medicare/Coding/MedHCPCSGenInfo/index} diagnosis and procedure code grouping. Floating point values were quantized based on the statistical distributions of each such variable. ICD codes for diagnoses and HCPCS procedure codes were rolled into their corresponding chapter-level categories. This lightweight feature engineering procedure follows CMS coding rules and guidelines and was co-designed with clinical health experts.

As a result of the above transformations, we reduced the vocabulary from more than 10 million unique tokens to approximately 20 thousand, a comparable vocabulary size to many deep neural language models \cite{DBLP:journals/corr/abs-1902-10339}. While this limited the specificity of input data, it allowed the model to more accurately learn the relationship between diagnoses groups and readmission outcomes. 

\subsection{Deep Learning Modeling}

\begin{figure*}
    \centering
    \includegraphics[width=\textwidth, keepaspectratio]{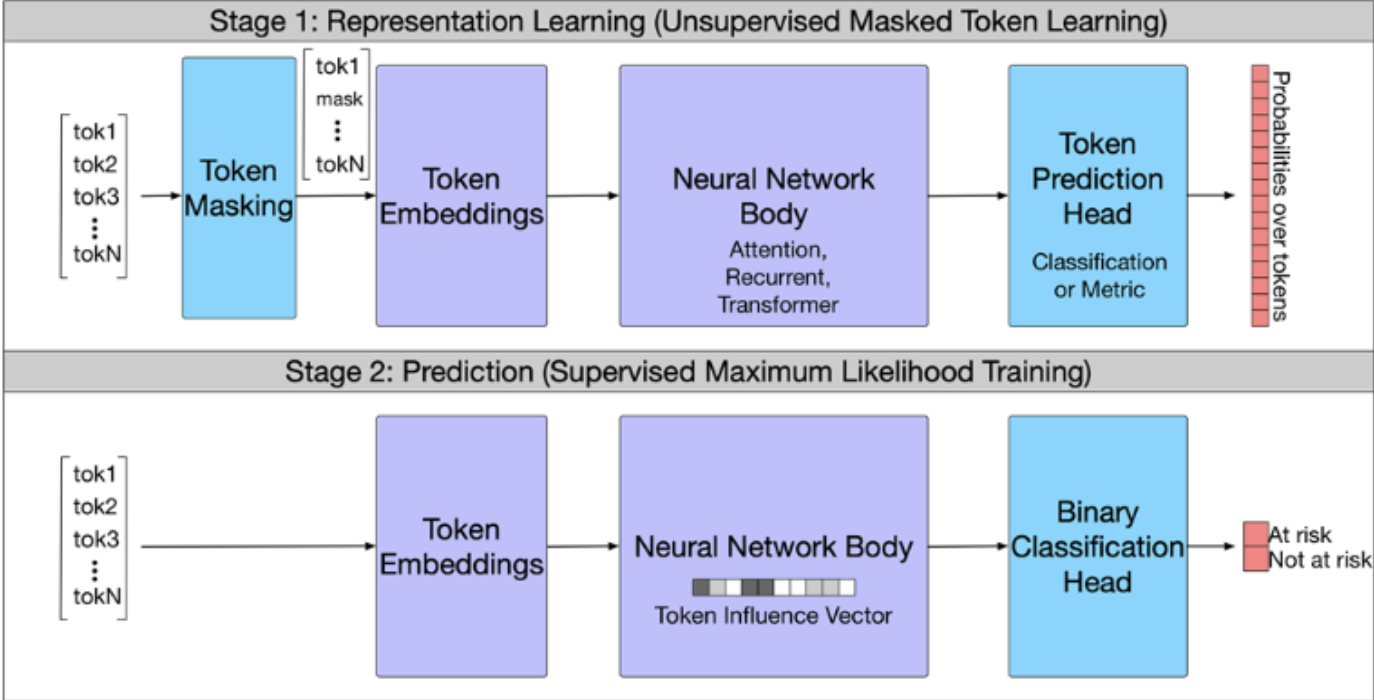}
    \caption{The two-stage deep learning modeling framework consisting of the unsupervised training to learn contextual relationships and the supervised model training stage to make unplanned readmission predictions.}
    \label{fig:dl}
\end{figure*}

As illustrated by Fig.~\ref{fig:dl}, our deep learning solution consists of two training stages: (1) transform categorical claims data such as claim code sets, e.g. diagnosis/procedure, and demographic information into numeric representations to learn contextual relationships amongst them; (2) finetune the model from (1) to make accurate predictions on the risk of unplanned readmissions. 

The purpose of stage 1 is to learn numeric representations of medical concepts. The data transformation step, hereon referred to as \textit{medical concept embedding}, is necessary to represent complex claims data in a format conducive for training deep learning models, preserving  the information in the data while uncovering implicit interrelations among predictor variables. Inspired by the popular idea of Masked Language Modeling (MLM) used to train state-of-the-art language models such as BERT \cite{DBLP:journals/corr/abs-1810-04805} and Transformer-XL \cite{DBLP:journals/corr/abs-1901-02860}, which both learn semantic and orthographic relationships between word tokens, we introduce the medical equivalent which captures the cross-domain, complex relationship of medical concepts which we refer to as Masked Token Learning (MTL). Compared with existing medical concept embedding techniques such as Med2Vec \cite{DBLP:journals/corr/ChoiBSCS16}, MTL learns co-occurrence behavior without context restrictions thereby capturing the additional structure found in longer claims. The MTL training task takes a vector of input tokens, e.g. $$[\text{tok}_1, \text{tok}_2, \dots, \text{tok}_N],$$ that may be any categorical, non-numeric data, masks a randomly selected token, and uses the context of the remaining tokens to predict it. The network architecture for this unsupervised training stage includes a token embedding layer, a network body, and a token prediction head. MTL-style training procedures paired with this network architecture allow for more informative features to be extracted, as compared with shallower networks like Med2Vec, without sacrificing explainability. The discussion on explainabilty is in Section~\ref{sec:explain}. In addition, to acknowledge the temporal relationships implicit to medical concepts, we include a time-aware attention mechanism in the model architecture. For the CMS AI Challenge prediction task on unplanned hospital 30 day readmissions, we used a BERT-like model architecture for the MTL task which is discussed in detail in Section \ref{sec: eval}.

The supervised training stage generates predictions of beneficiary risk for unplanned readmission by fine-tuning the models learned in the first, unsupervised stage. Specifically, the MTL network head is exchanged for one designed for binary classification on token embeddings.  Using the classical logistic regression approach, the trained model generates predictions for unplanned readmission associated  with a given beneficiary.

\subsection{AI Explainability}\label{sec:explain}

Although deep learning and language models can aid in predictive performance, in order to garner end user buy-in among subject matter experts (SMEs) and decision makers it is necessary to provide explainable predictions. This enables knowledge building and further examination of model outputs by trained medical professionals. A number of studies have leveraged dimensional reduction techniques, such as t-distributed stochastic neighbor embeddings (t-SNE), to construct clusters that may lend itself to an interpretation of learned representations \cite{Xu2018,Choi2017,Ma2018,Tran2015,Zhang2019,Si2019,Lei2019}. Other papers have tackled explainability through using visualization of features and feature importances \cite{Cho2014,Baytas2017,Zhang2018}. Although these were important steps to understanding model output, there are notable limitations to using dimension reduction techniques and feature importances. In the case of dimension reduction techniques, they only provide a suggestive view and require an element of human interpretation to give meaning to clusters. The limitation of feature importances lie in that they are typically derived from models that learn population-level importances versus individual-level importances. In our case, individual-level feature importances are preferred as we would like to explain predictions for each beneficiary.

A benefit of our approach, modeling medical claims with a Transformer-based model, is that we can leverage the attention mechanism to reveal internal probability vectors over which claim tokens (i.e.,variables) are influential to the final classification. There have been a few health papers using attention mechanisms for interpretability examining issues such as patient classification using clinical notes \cite{Kemp2019}, patient risk prediction using EHRs \cite{Ma2018a}, diagnosis prediction using EHRs \cite{Ma2017,Choi2016}, and predicting ICU readmission risk using EHRs \cite{Barbieri2020}. To the best of our knowledge, we contribute to the first case of using attention mechanisms in the context of a large language model to explain patient readmission using Medicare claims data.

\section{Performance Evaluation} \label{sec: eval}

The proposed framework was piloted through a prediction task on unplanned hospital readmissions within 30 days of discharge for Medicare beneficiaries. Following the approach outlined in Section \ref{sec:approach}, we used a custom-configured BERT architecture and the AdamW optimizer for training tasks with various dataset sizes. This configuration and architecture provided the necessary balance between AUC and training time, allowing for quick model iterations.

\subsection{Model Training} 

We performed several experiments with subsets of 200K and 600K inpatient labeled events for 50K and 150K unique beneficiaries, respectively. Both datasets had an approximately 50/50 split between readmission and not-readmission labels. We then scaled to the entire subset of inpatient claims between 2010 and 2011 to build approximately 1.2M unique inpatient labeled events from approximately 600K unique beneficiaries, with a 3:1 split between claims labeled with a 0 and claims labeled with a 1. Table \ref{table:model_eval1} shows markedly improved AUC and recall as the size of the training set increased, demonstrating the model's capacity to learn from larger and richer datasets. 

\begin{table}[!h]
\begin{tabular}{@{}clcc@{}}
\toprule
\textbf{Dataset Size} & \multicolumn{1}{c}{\textbf{AUC}} & \textbf{Recall} \\ \midrule
200K & 0.72 & 0.33 \\
600K & 0.72 & 0.61\\
1.2M & 0.91 & 0.91 \\
\bottomrule
\end{tabular}
\caption{AUC and recall improvements as the size of the training set increased demonstrates the model's capacity to learn from larger claims datasets even in the face of 3:1 label skew and patient multi-sampling. This is important given the model's goal to perform on all U.S. Medicare recipients.}
\label{table:model_eval1}
\end{table}

Additionally, initial experimental results indicated the semi-structured nature of ordering tabular claims data in sequential form was simplistic for the model's size and complexity, resulting in over-fitting during MTL training. We then investigated smaller architectures, such as Distil-BERT, and found that a reduced BERT model architecture of two hidden layers, two attention layers, and a hidden dimension size of 512 achieved the best-fit learning curves in our experiments.

Training was conducted with a data split of 80\% (train), 10\% (validate), 10\% (test) to monitor early stop. Since our labeling approach allowed for multi-sampling of a beneficiary's history, we ensured there would be no beneficiary overlap between train/validate/test. As specified in \cite{Devlin2019}, 15\% of tokens are changed to either a mask or replaced with another random token during the MTL training phase. We utilized an AdamW optimizer \cite{loshchilov2019decoupled} with cosine scheduling \cite{loshchilov2017sgdr} in both the MTL and classification training. 

\subsection{Model Performance}

We achieved an AUC of 0.91 and recall of 0.91. Recall is particularly important since greater health and financial risks are associated with false negative predictions, i.e. missing a beneficiary's readmission. This performance significantly exceeds prior work in patient readmission prediction whose AUC's, detailed in Table \ref{table:performancecomparison}, range the 60's and 70's and often relied on the richer and more complete patient histories found in electronic health records (EHR).

\begin{table}[!h]
\begin{tabular}{@{}clcc@{}}
\toprule
\textbf{Paper} & \multicolumn{1}{c}{\textbf{Approach}} & \textbf{Horizon} & \textbf{AUC} \\ \midrule
\cite{Yang2017} & Tensor decomposition & 1 year  & 0.68-0.78 \\
\cite{Zhang2018} & EHR embedding & 6 months & 0.79 \\
\cite{Brungger2019} & \begin{tabular}[c]{@{}l@{}}Logistic regression +\\ medical claims (Switzerland)\end{tabular} & 30 days & 0.60-0.61 \\
\cite{Xiao2018} & TopicRNN on EHR & 30 days & 0.61 \\
\cite{Wang2018} & CNN + MLP on EMR & 30 days & 0.70 \\
\cite{Rajkomar2018} & RNN on EHR & 30 days & 0.75-0.76 \\
\cite{Min2019} & \begin{tabular}[c]{@{}l@{}}Mix of ML,DL, embeddings, \\ \& attention models\end{tabular} & 30 days & 0.50-0.65 \\
\textit{This Work} & BERT + medical claims & 30 days & \textbf{0.91} \\ 
\bottomrule
\end{tabular}
\caption{Comparison of model performance across the patient readmission literature. Previous papers vary by approach and horizon. Our work shows a significant increase in performance over these approaches based on AUC.}
\label{table:performancecomparison}
\end{table}

To account for the inherent label skewness in the final 1.2M dataset, we tuned the classification threshold $\alpha$ as follows:
\begin{equation}
    \hat{\alpha} = \arg\max_{\alpha} G-Mean(\alpha)
\end{equation}

where
\begin{equation}
    G-Mean(\alpha)=\sqrt{Sensitivity(\alpha) \cdot Specificity(\alpha)}
\end{equation}

and $Sensitivity(\alpha)$ and $Specificity(\alpha)$ are the usual calculations where the classification rule depends on $\alpha$ such that the predicted label $\hat{y_{i}}$ is defined as 
\begin{equation}
    \hat{y_{i}} = \mathbbm{1}(Pr(\text{30-day readmission}) > \alpha)
\end{equation}

and $\mathbbm{1}(\cdot)$ represents the indicator function where if the argument in parentheses is true it assigns a value of 1 and 0 otherwise.

The selected threshold $\hat{\alpha}$ trades off between specificity and sensitivity or penalties associated with false positives and false negatives, accordingly.

\subsection{Attention Vectors for Explainability}

\begin{figure*}[!h]
    \centering
    \includegraphics[width=\textwidth, keepaspectratio]{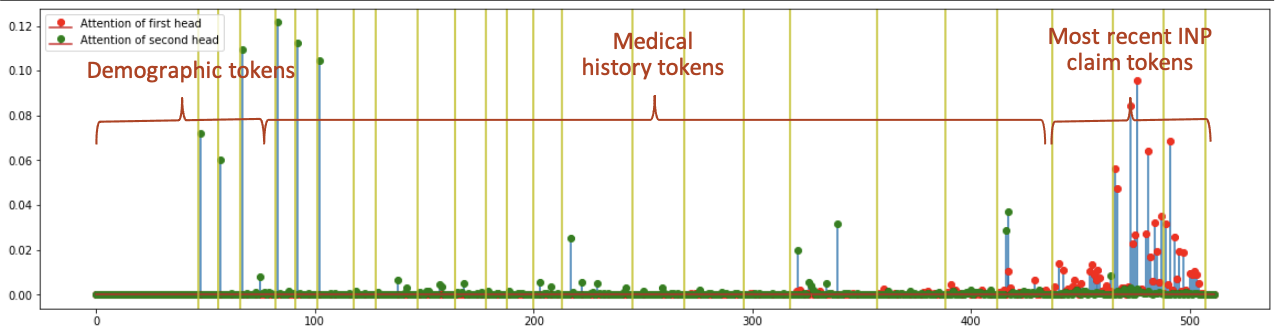}
    \caption{Example attention vector output showing attention probabilities associated with each token input across the two attention layers. Red markers indicate attention from the first head and blue markers indicate attention from the second head. The model’s focus on demographic information and recent claims follow our healthcare SME evaluators expectations.}\label{fig:attention}
\end{figure*}

 Using the trained Transformer's attention outputs, we analyzed which claim tokens or variables were important to the model prediction. Figure \ref{fig:attention} is an example attention output, a vector of probabilities corresponding the position of the input tokens. Therefore, The vector begins with the attention on demographic and Social Determinants of Health information, and then lists the tokens from the beneficiary's 3 month claim history from oldest to most recent.  The model’s focus on demographic information and recent claims indicates that the model learns mostly from the social determinants of health and recent history. This is consistent with the expectations of the healthcare SME evaluators we had review randomly selected samples. Detailed and specific information, such as attention, is vital to explaining results to health care providers and stakeholders, and can inform beneficiary-specific medical interventions. Moreover, it is useful information to model bias detection and mitigation as well as many of the primary challenges we believe need to be addressed associated with model deployment and operationalization.

\section{Considerations for Operationalization} \label{sec:op}

\subsection{Deployment}

We recognize that our model is only useful if it can enable health service providers to make more informed patient and population intervention decisions.  We believe that among the primary challenges to this goal are the following:
\begin{itemize}
    \item Healthcare information dissemination channels are complex and often involve proprietary data storage and transfer methods
    \item Healthcare providers must efficiently synthesize a variety of disparate data points associated with a patient in order to make complex intervention decisions.
\end{itemize}

In order to begin to address the first challenge, we recommend a deployment architecture that relies heavily on emerging and well-known standards-compliant channels.  For example, we believe that a Health Level 7-compliant\footnote{See \href{https://www.hl7.org/}{\textit{hl7.org}}} ADT (admissions, discharges, and transfers) feed available through Health Information Exchanges pushed through Direct Secure Messages\footnote{See \href{https://wiki.directproject.org/}{\textit{wiki.directproject.org}}} could provide patient-specific or even population-level reports to member healthcare systems.  We also expect that a \textit{SMART on FHIR}\footnote{See \href{https://smarthealthit.org}{\textit{smarthealthit.org}}} application programming interface could deliver information directly into an electronic health record notification panel using the clinical decision support (CDS) Hook standard\footnote{See \href{https://cds-hooks.org/}{\textit{cds-hooks.org}}}.  Ultimately, bringing the results to the health service provider within the context where they operate will be necessary to ensure front-line clinicians are able to effectively consume the information our model provides.

\begin{figure}[!h]
    \centering
    \includegraphics[width=\columnwidth, keepaspectratio]{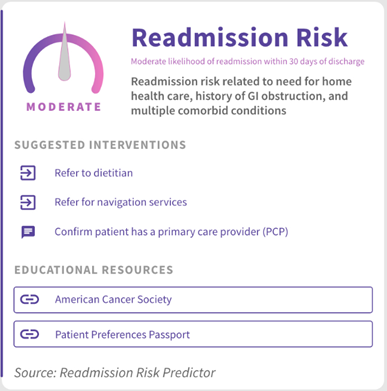}
    \caption{Our prototype notification card would be triggered when a patient's chart is opened within an EHR through CDS Hooks. The \textit{Suggested Interventions} are translated from the most important tokens in the attention vector. All the data in this card is representative, but not attributable to any CMS beneficiary.}
    \label{fig:Card_Example}
\end{figure}

\begin{figure}[!h]
    \centering
    \includegraphics[width=\columnwidth, keepaspectratio]{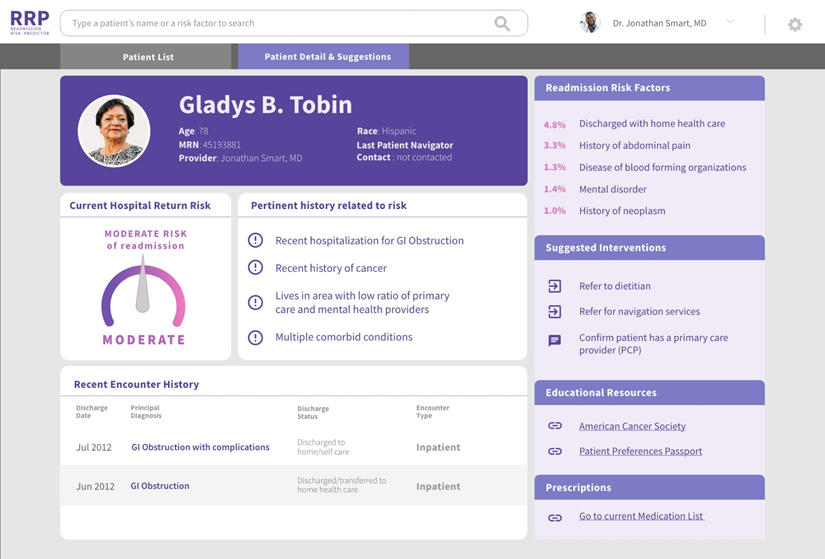}
    \caption{\textit{Readmission Risk Factors} and \textit{Suggested Interventions} translated from the most important tokens in the attention vector. Accessed from the EHR (\textit{e.g.} through the Notification Card shown in Fig. \ref{fig:Card_Example}), providing additional risk details and contextual information about the patient's risk for readmission.  It also displays the approximate amount that each risk factor contributes to the overall risk. All the data in this graphic is representative, but not attributable to any CMS beneficiary.}
    \label{fig:UI_Example}
\end{figure}

In order to address the challenge of getting the right information to the provider at the right time in a manner that does not add to alert-fatigue\cite{alertfatigue}, we designed a series of prototype user interfaces with the help of subject matter experts in the fields of medicine and patient care.  In order to populate the information within these user interfaces, we first translated the ranked input tokens from the attention vector into beneficiary-specific intervention recommendations that could maximize individual and network-wide risk reduction. We then worked with human-centered design experts to organize the resulting information in a variety of ways to fit into existing electronic health record systems. For example, Fig. \ref{fig:Card_Example} shows a small card that could pop into the notification panel of an electronic health record system within the context of the patient panel.  Furthermore, if a clinician wanted a more in-depth exploration of readmission risk, Fig. \ref{fig:UI_Example} shows what a full application would look like, including the display of other contextual data (co-morbidities, patient history, Social Determinants of Health).  Ultimately, ensuring that AI Explainability takes a variety of consumption paradigms into account is vital to translating the data a model produces into effective action.

\subsection{Generalizability}

A fundamental challenge associated with health applications pertains to the diversity and non-stationarity found in real-world medical records, e.g. there are no COVID-19 records in 2018. As expected, applying our health risk model, which was trained on 2008-2011, to the medical records from 2012 yielded a lower AUC of 0.81. A major component of our solution involves the understanding that continued use of the model in practice would require continuous retraining and monitoring for data and model drift.

\subsection{Bias}

Another important consideration for model deployment is bias. In part motivated by Senator Booker’s specific request to control and mitigate bias in models developed in the CMS AI Health Risk Challenge \footnote{https://www.booker.senate.gov/news/press/booker-wyden-demand-answers-on-biased-health-care-algorithms} and the general goal of developing ethical and equitable AI solutions, we investigated how sensitive our model predictions are to patient demographics. From the the model's attention mechanism, we observed that attention is largely paid to clinically relevant features as well as demographic information. Specifically, within the top 3 highest attention variables, "provider specialty" appears most frequently at 66\%, followed by "type of service" and "place of service" at 41\% and 37\%, respectively.

To further investigate, we measured the model's performance across race and gender subgroups and re-evaluated with a model trained with a modified dataset where altered or masked race- and gender-related variables. This analysis identified overpredictiton of readmission for certain underrepresented subgroups. While outside the scope of this paper, this finding warrants further study to identify sub-populations where readmission risk stratification is biased in order to more effectively reduce risk across the entire CMS population. 

\section{Conclusion}

The unplanned readmission of a patient is costly to both the CMS and the patient themself. Moreover, it potentially signals insufficient care or an inadequate assessment of the patients' future health risks. Encapsulated by the CMS AI Health Outcomes Challenge, this paper proposed a unique approach to addressing this problem by predicting the likelihood of an unplanned readmission in a 30-day time horizon through leveraging a Transformer-based model trained on medical claims records. To support this, we demonstrated a novel data pre-processing pipeline and showed a significant performance increase over previous approaches to the problem which leverage various types of machine learning and deep learning models.

This paper discussed the explainability afforded by the use of attention mechanisms as well as concerns regarding biases, both important aspects of using deep learning methods for prediction tasks in healthcare applications. To make the paper more accessible to practitioners, we also discussed the deployment of the approach making use of existing workflows and translated our model results into an intuitive user interface for health professionals.

\section*{Acknowledgements}
The authors would like to acknowledge the technical contributions from Matt Anderson, Sylvia Chin, Greg Miernicki, and Mortiz Steller during development, Edward Raff for manuscript review and editing. This work benefited from the medical expertise of Barb Doyle, Wendy Watson, Amy Kress Youssef, and Sam Zhang. Furthermore, the deployment paradigm and UI design benefited from discussions with medical experts from The MedStar Institute for Innovation, Louisiana Public Health Initiative, and Planetree. Finally, we appreciated the opportunity to participate in the CMS AI Health Outcomes Challenge and the privilege to work on this problem and share our approach.

\section*{Disclaimer}
The views expressed in this paper reflect those of the authors and not of Booz Allen Hamilton, Centers for Medicare \& Medicaid Services, or the U.S. federal government.

\bibliographystyle{ACM-Reference-Format}
\bibliography{references.bib}


\begin{thebibliography}{54}


\ifx \showCODEN    \undefined \def \showCODEN     #1{\unskip}     \fi
\ifx \showDOI      \undefined \def \showDOI       #1{#1}\fi
\ifx \showISBNx    \undefined \def \showISBNx     #1{\unskip}     \fi
\ifx \showISBNxiii \undefined \def \showISBNxiii  #1{\unskip}     \fi
\ifx \showISSN     \undefined \def \showISSN      #1{\unskip}     \fi
\ifx \showLCCN     \undefined \def \showLCCN      #1{\unskip}     \fi
\ifx \shownote     \undefined \def \shownote      #1{#1}          \fi
\ifx \showarticletitle \undefined \def \showarticletitle #1{#1}   \fi
\ifx \showURL      \undefined \def \showURL       {\relax}        \fi
\providecommand\bibfield[2]{#2}
\providecommand\bibinfo[2]{#2}
\providecommand\natexlab[1]{#1}
\providecommand\showeprint[2][]{arXiv:#2}

\bibitem[\protect\citeauthoryear{A., H., M., M., U., and J.}{A.
  et~al\mbox{.}}{2015}]%
        {A.2015}
\bibfield{author}{\bibinfo{person}{Alassaad A.}, \bibinfo{person}{Melhus H.},
  \bibinfo{person}{Hammarlund-Udenaes M.}, \bibinfo{person}{Bertilsson M.},
  \bibinfo{person}{Gillespie U.}, {and} \bibinfo{person}{Sundstr{\"{o}}m J.}}
  \bibinfo{year}{2015}\natexlab{}.
\newblock \showarticletitle{{A Tool for prediction of risk of rehospitalisation
  and mortality in the hospitalised elderly: Secondary analysis of clinical
  trial data}}.
\newblock \bibinfo{journal}{\emph{BMJ Open}} (\bibinfo{year}{2015}).
\newblock
\showISSN{2044-6055}


\bibitem[\protect\citeauthoryear{Alsentzer, Murphy, Boag, Weng, Jin, Naumann,
  and McDermott}{Alsentzer et~al\mbox{.}}{2019}]%
        {Alsentzer2019}
\bibfield{author}{\bibinfo{person}{Emily Alsentzer}, \bibinfo{person}{John~R.
  Murphy}, \bibinfo{person}{Willie Boag}, \bibinfo{person}{Wei~Hung Weng},
  \bibinfo{person}{Di Jin}, \bibinfo{person}{Tristan Naumann}, {and}
  \bibinfo{person}{Matthew~B.A. McDermott}.} \bibinfo{year}{2019}\natexlab{}.
\newblock \bibinfo{title}{{Publicly available clinical BERT embeddings}}.
\newblock
\newblock
\showISSN{23318422}
\showeprint[arxiv]{1904.03323}


\bibitem[\protect\citeauthoryear{Ancker, Edwards, Nosal, Hauser, Mauer,
  Kaushal, and Investigators}{Ancker et~al\mbox{.}}{2017}]%
        {alertfatigue}
\bibfield{author}{\bibinfo{person}{Jessica Ancker}, \bibinfo{person}{Alison
  Edwards}, \bibinfo{person}{Sarah Nosal}, \bibinfo{person}{Diane Hauser},
  \bibinfo{person}{Elizabeth Mauer}, \bibinfo{person}{Rainu Kaushal}, {and}
  \bibinfo{person}{with Investigators}.} \bibinfo{year}{2017}\natexlab{}.
\newblock \showarticletitle{Effects of workload, work complexity, and repeated
  alerts on alert fatigue in a clinical decision support system}.
\newblock \bibinfo{journal}{\emph{BMC Medical Informatics and Decision Making}}
   \bibinfo{volume}{17} (\bibinfo{date}{04} \bibinfo{year}{2017}),
  \bibinfo{pages}{36}.
\newblock
\urldef\tempurl%
\url{https://doi.org/10.1186/s12911-017-0430-8}
\showDOI{\tempurl}


\bibitem[\protect\citeauthoryear{Artetxe, Beristain, and Gra{\~{n}}a}{Artetxe
  et~al\mbox{.}}{2018}]%
        {Artetxe2018}
\bibfield{author}{\bibinfo{person}{Arkaitz Artetxe}, \bibinfo{person}{Andoni
  Beristain}, {and} \bibinfo{person}{Manuel Gra{\~{n}}a}.}
  \bibinfo{year}{2018}\natexlab{}.
\newblock \showarticletitle{{Predictive models for hospital readmission risk: A
  systematic review of methods}}.
\newblock \bibinfo{journal}{\emph{Computer Methods and Programs in
  Biomedicine}}  \bibinfo{volume}{164} (\bibinfo{year}{2018}),
  \bibinfo{pages}{49--64}.
\newblock
\showISSN{18727565}
\urldef\tempurl%
\url{https://doi.org/10.1016/j.cmpb.2018.06.006}
\showDOI{\tempurl}


\bibitem[\protect\citeauthoryear{Barbieri, Kemp, Perez-Concha, Kotwal,
  Gallagher, Ritchie, and Jorm}{Barbieri et~al\mbox{.}}{2020}]%
        {Barbieri2020}
\bibfield{author}{\bibinfo{person}{Sebastiano Barbieri}, \bibinfo{person}{James
  Kemp}, \bibinfo{person}{Oscar Perez-Concha}, \bibinfo{person}{Sradha Kotwal},
  \bibinfo{person}{Martin Gallagher}, \bibinfo{person}{Angus Ritchie}, {and}
  \bibinfo{person}{Louisa Jorm}.} \bibinfo{year}{2020}\natexlab{}.
\newblock \showarticletitle{{Benchmarking Deep Learning Architectures for
  Predicting Readmission to the ICU and Describing Patients-at-Risk}}.
\newblock \bibinfo{journal}{\emph{Scientific Reports}} \bibinfo{volume}{10},
  \bibinfo{number}{1} (\bibinfo{year}{2020}), \bibinfo{pages}{1--10}.
\newblock
\showISSN{20452322}
\urldef\tempurl%
\url{https://doi.org/10.1038/s41598-020-58053-z}
\showDOI{\tempurl}
\showeprint[arxiv]{1905.08547}


\bibitem[\protect\citeauthoryear{Baytas, Xiao, Zhang, Wang, Jain, and
  Zhou}{Baytas et~al\mbox{.}}{2017}]%
        {Baytas2017}
\bibfield{author}{\bibinfo{person}{Inci~M. Baytas}, \bibinfo{person}{Cao Xiao},
  \bibinfo{person}{Xi Zhang}, \bibinfo{person}{Fei Wang},
  \bibinfo{person}{Anil~K. Jain}, {and} \bibinfo{person}{Jiayu Zhou}.}
  \bibinfo{year}{2017}\natexlab{}.
\newblock \showarticletitle{{Patient subtyping via time-aware LSTM networks}}.
\newblock \bibinfo{journal}{\emph{Proceedings of the ACM SIGKDD International
  Conference on Knowledge Discovery and Data Mining}}  \bibinfo{volume}{Part
  F1296} (\bibinfo{year}{2017}), \bibinfo{pages}{65--74}.
\newblock
\showISBNx{9781450348874}
\urldef\tempurl%
\url{https://doi.org/10.1145/3097983.3097997}
\showDOI{\tempurl}


\bibitem[\protect\citeauthoryear{Br{\"{u}}ngger and Blozik}{Br{\"{u}}ngger and
  Blozik}{2019}]%
        {Brungger2019}
\bibfield{author}{\bibinfo{person}{Beat Br{\"{u}}ngger} {and}
  \bibinfo{person}{Eva Blozik}.} \bibinfo{year}{2019}\natexlab{}.
\newblock \showarticletitle{{Hospital readmission risk prediction based on
  claims data available at admission: A pilot study in Switzerland}}.
\newblock \bibinfo{journal}{\emph{BMJ Open}} (\bibinfo{year}{2019}).
\newblock
\showISSN{20446055}
\urldef\tempurl%
\url{https://doi.org/10.1136/bmjopen-2018-028409}
\showDOI{\tempurl}


\bibitem[\protect\citeauthoryear{Chen, Su, Shen, Chen, Yan, and Wang}{Chen
  et~al\mbox{.}}{2019}]%
        {DBLP:journals/corr/abs-1902-10339}
\bibfield{author}{\bibinfo{person}{Wenhu Chen}, \bibinfo{person}{Yu Su},
  \bibinfo{person}{Yilin Shen}, \bibinfo{person}{Zhiyu Chen},
  \bibinfo{person}{Xifeng Yan}, {and} \bibinfo{person}{William~Yang Wang}.}
  \bibinfo{year}{2019}\natexlab{}.
\newblock \showarticletitle{How Large a Vocabulary Does Text Classification
  Need? {A} Variational Approach to Vocabulary Selection}.
\newblock \bibinfo{journal}{\emph{CoRR}}  \bibinfo{volume}{abs/1902.10339}
  (\bibinfo{year}{2019}).
\newblock
\showeprint[arxiv]{1902.10339}
\urldef\tempurl%
\url{http://arxiv.org/abs/1902.10339}
\showURL{%
\tempurl}


\bibitem[\protect\citeauthoryear{Cho, van Merrienboer, Bahdanau, and
  Bengio}{Cho et~al\mbox{.}}{2014}]%
        {Cho2014}
\bibfield{author}{\bibinfo{person}{Kyunghyun Cho}, \bibinfo{person}{Bart van
  Merrienboer}, \bibinfo{person}{Dzmitry Bahdanau}, {and}
  \bibinfo{person}{Yoshua Bengio}.} \bibinfo{year}{2014}\natexlab{}.
\newblock \showarticletitle{{On the Properties of Neural Machine Translation:
  Encoder-Decoder Approaches}}.
\newblock \bibinfo{journal}{\emph{arXiv}} (\bibinfo{date}{sep}
  \bibinfo{year}{2014}), \bibinfo{pages}{103--111}.
\newblock
\urldef\tempurl%
\url{https://doi.org/10.3115/v1/w14-4012}
\showDOI{\tempurl}
\showeprint[arxiv]{1409.1259}


\bibitem[\protect\citeauthoryear{Choi, Bahadori, Kulas, Schuetz, Stewart, and
  Sun}{Choi et~al\mbox{.}}{2016a}]%
        {Choi2016}
\bibfield{author}{\bibinfo{person}{Edward Choi}, \bibinfo{person}{Mohammad~Taha
  Bahadori}, \bibinfo{person}{Joshua~A. Kulas}, \bibinfo{person}{Andy Schuetz},
  \bibinfo{person}{Walter~F. Stewart}, {and} \bibinfo{person}{Jimeng Sun}.}
  \bibinfo{year}{2016}\natexlab{a}.
\newblock \showarticletitle{{RETAIN: An interpretable predictive model for
  healthcare using reverse time attention mechanism}}.
\newblock \bibinfo{journal}{\emph{Advances in Neural Information Processing
  Systems}} \bibinfo{number}{Nips} (\bibinfo{year}{2016}),
  \bibinfo{pages}{3512--3520}.
\newblock
\showISSN{10495258}
\showeprint[arxiv]{1608.05745}


\bibitem[\protect\citeauthoryear{Choi, Bahadori, Schuetz, Stewart, and
  Sun}{Choi et~al\mbox{.}}{2016b}]%
        {Choi2016a}
\bibfield{author}{\bibinfo{person}{Edward Choi}, \bibinfo{person}{Mohammad~Taha
  Bahadori}, \bibinfo{person}{Andy Schuetz}, \bibinfo{person}{Walter~F
  Stewart}, {and} \bibinfo{person}{Jimeng Sun}.}
  \bibinfo{year}{2016}\natexlab{b}.
\newblock \showarticletitle{{Doctor AI: Predicting Clinical Events via
  Recurrent Neural Networks.}}
\newblock \bibinfo{journal}{\emph{JMLR workshop and conference proceedings}}
  \bibinfo{volume}{56} (\bibinfo{year}{2016}), \bibinfo{pages}{301--318}.
\newblock
\showISSN{1938-7288}
\showeprint[arxiv]{1511.05942}
\urldef\tempurl%
\url{http://www.ncbi.nlm.nih.gov/pubmed/28286600{\%}0Ahttp://www.pubmedcentral.nih.gov/articlerender.fcgi?artid=PMC5341604}
\showURL{%
\tempurl}


\bibitem[\protect\citeauthoryear{Choi, Bahadori, Searles, Coffey, and Sun}{Choi
  et~al\mbox{.}}{2016c}]%
        {DBLP:journals/corr/ChoiBSCS16}
\bibfield{author}{\bibinfo{person}{Edward Choi}, \bibinfo{person}{Mohammad~Taha
  Bahadori}, \bibinfo{person}{Elizabeth Searles}, \bibinfo{person}{Catherine
  Coffey}, {and} \bibinfo{person}{Jimeng Sun}.}
  \bibinfo{year}{2016}\natexlab{c}.
\newblock \showarticletitle{Multi-layer Representation Learning for Medical
  Concepts}.
\newblock \bibinfo{journal}{\emph{CoRR}}  \bibinfo{volume}{abs/1602.05568}
  (\bibinfo{year}{2016}).
\newblock
\showeprint[arxiv]{1602.05568}
\urldef\tempurl%
\url{http://arxiv.org/abs/1602.05568}
\showURL{%
\tempurl}


\bibitem[\protect\citeauthoryear{Choi, Bahadori, Song, Stewart, and Sun}{Choi
  et~al\mbox{.}}{2017}]%
        {Choi2017}
\bibfield{author}{\bibinfo{person}{Edward Choi}, \bibinfo{person}{Mohammad~Taha
  Bahadori}, \bibinfo{person}{Le Song}, \bibinfo{person}{Walter~F. Stewart},
  {and} \bibinfo{person}{Jimeng Sun}.} \bibinfo{year}{2017}\natexlab{}.
\newblock \showarticletitle{{GRAM: Graph-based attention model for healthcare
  representation learning}}.
\newblock \bibinfo{journal}{\emph{Proceedings of the ACM SIGKDD International
  Conference on Knowledge Discovery and Data Mining}}  \bibinfo{volume}{Part
  F1296} (\bibinfo{year}{2017}), \bibinfo{pages}{787--795}.
\newblock
\showISBNx{9781450348874}
\urldef\tempurl%
\url{https://doi.org/10.1145/3097983.3098126}
\showDOI{\tempurl}
\showeprint[arxiv]{1611.07012}


\bibitem[\protect\citeauthoryear{Corrigan and Martin}{Corrigan and
  Martin}{1992}]%
        {Corrigan1992}
\bibfield{author}{\bibinfo{person}{J~M Corrigan} {and} \bibinfo{person}{J~B
  Martin}.} \bibinfo{year}{1992}\natexlab{}.
\newblock \showarticletitle{{Identification of factors associated with hospital
  readmission and development of a predictive model.}}
\newblock \bibinfo{journal}{\emph{Health services research}}
  (\bibinfo{year}{1992}).
\newblock
\showISSN{0017-9124}


\bibitem[\protect\citeauthoryear{Cox}{Cox}{1972}]%
        {Cox1972}
\bibfield{author}{\bibinfo{person}{D.~R. Cox}.}
  \bibinfo{year}{1972}\natexlab{}.
\newblock \showarticletitle{{Regression Models and Life-Tables}}.
\newblock \bibinfo{journal}{\emph{Journal of the Royal Statistical Society:
  Series B (Methodological)}} (\bibinfo{year}{1972}).
\newblock
\urldef\tempurl%
\url{https://doi.org/10.1111/j.2517-6161.1972.tb00899.x}
\showDOI{\tempurl}


\bibitem[\protect\citeauthoryear{Dai, Yang, Yang, Carbonell, Le, and
  Salakhutdinov}{Dai et~al\mbox{.}}{2019}]%
        {DBLP:journals/corr/abs-1901-02860}
\bibfield{author}{\bibinfo{person}{Zihang Dai}, \bibinfo{person}{Zhilin Yang},
  \bibinfo{person}{Yiming Yang}, \bibinfo{person}{Jaime~G. Carbonell},
  \bibinfo{person}{Quoc~V. Le}, {and} \bibinfo{person}{Ruslan Salakhutdinov}.}
  \bibinfo{year}{2019}\natexlab{}.
\newblock \showarticletitle{Transformer-XL: Attentive Language Models Beyond a
  Fixed-Length Context}.
\newblock \bibinfo{journal}{\emph{CoRR}}  \bibinfo{volume}{abs/1901.02860}
  (\bibinfo{year}{2019}).
\newblock
\showeprint[arxiv]{1901.02860}
\urldef\tempurl%
\url{http://arxiv.org/abs/1901.02860}
\showURL{%
\tempurl}


\bibitem[\protect\citeauthoryear{Devlin, Chang, Lee, and Toutanova}{Devlin
  et~al\mbox{.}}{2018}]%
        {DBLP:journals/corr/abs-1810-04805}
\bibfield{author}{\bibinfo{person}{Jacob Devlin}, \bibinfo{person}{Ming{-}Wei
  Chang}, \bibinfo{person}{Kenton Lee}, {and} \bibinfo{person}{Kristina
  Toutanova}.} \bibinfo{year}{2018}\natexlab{}.
\newblock \showarticletitle{{BERT:} Pre-training of Deep Bidirectional
  Transformers for Language Understanding}.
\newblock \bibinfo{journal}{\emph{CoRR}}  \bibinfo{volume}{abs/1810.04805}
  (\bibinfo{year}{2018}).
\newblock
\showeprint[arxiv]{1810.04805}
\urldef\tempurl%
\url{http://arxiv.org/abs/1810.04805}
\showURL{%
\tempurl}


\bibitem[\protect\citeauthoryear{Devlin, Chang, Lee, and Toutanova}{Devlin
  et~al\mbox{.}}{2019}]%
        {Devlin2019}
\bibfield{author}{\bibinfo{person}{Jacob Devlin}, \bibinfo{person}{Ming~Wei
  Chang}, \bibinfo{person}{Kenton Lee}, {and} \bibinfo{person}{Kristina
  Toutanova}.} \bibinfo{year}{2019}\natexlab{}.
\newblock \showarticletitle{{BERT: Pre-training of deep bidirectional
  transformers for language understanding}}. In \bibinfo{booktitle}{\emph{NAACL
  HLT 2019 - 2019 Conference of the North American Chapter of the Association
  for Computational Linguistics: Human Language Technologies - Proceedings of
  the Conference}}.
\newblock
\showISBNx{9781950737130}
\showeprint[arxiv]{1810.04805}


\bibitem[\protect\citeauthoryear{Hao, Wang, Jin, Shin, Zhu, Huang, Zheng, Luo,
  Hu, Fu, Dai, Wang, Culver, Alfreds, Rogow, Stearns, Sylvester, Widen, and
  Ling}{Hao et~al\mbox{.}}{2015}]%
        {Hao2015}
\bibfield{author}{\bibinfo{person}{Shiying Hao}, \bibinfo{person}{Yue Wang},
  \bibinfo{person}{Bo Jin}, \bibinfo{person}{Andrew~Young Shin},
  \bibinfo{person}{Chunqing Zhu}, \bibinfo{person}{Min Huang},
  \bibinfo{person}{Le Zheng}, \bibinfo{person}{Jin Luo},
  \bibinfo{person}{Zhongkai Hu}, \bibinfo{person}{Changlin Fu},
  \bibinfo{person}{Dorothy Dai}, \bibinfo{person}{Yicheng Wang},
  \bibinfo{person}{Devore~S. Culver}, \bibinfo{person}{Shaun~T. Alfreds},
  \bibinfo{person}{Todd Rogow}, \bibinfo{person}{Frank Stearns},
  \bibinfo{person}{Karl~G. Sylvester}, \bibinfo{person}{Eric Widen}, {and}
  \bibinfo{person}{Xuefeng~B. Ling}.} \bibinfo{year}{2015}\natexlab{}.
\newblock \showarticletitle{{Development, validation and deployment of a real
  time 30 day hospital readmission risk assessment tool in the Maine healthcare
  information exchange}}.
\newblock \bibinfo{journal}{\emph{PLoS ONE}} (\bibinfo{year}{2015}).
\newblock
\showISSN{19326203}
\urldef\tempurl%
\url{https://doi.org/10.1371/journal.pone.0140271}
\showDOI{\tempurl}


\bibitem[\protect\citeauthoryear{Ishwaran, Kogalur, Blackstone, and
  Lauer}{Ishwaran et~al\mbox{.}}{2008}]%
        {Ishwaran2008}
\bibfield{author}{\bibinfo{person}{Hemant Ishwaran}, \bibinfo{person}{Udaya~B.
  Kogalur}, \bibinfo{person}{Eugene~H. Blackstone}, {and}
  \bibinfo{person}{Michael~S. Lauer}.} \bibinfo{year}{2008}\natexlab{}.
\newblock \showarticletitle{{Random survival forests}}.
\newblock \bibinfo{journal}{\emph{Annals of Applied Statistics}}
  \bibinfo{volume}{2}, \bibinfo{number}{3} (\bibinfo{year}{2008}),
  \bibinfo{pages}{841--860}.
\newblock
\showISSN{19326157}
\urldef\tempurl%
\url{https://doi.org/10.1214/08-AOAS169}
\showDOI{\tempurl}
\showeprint[arxiv]{arXiv:0811.1645v1}


\bibitem[\protect\citeauthoryear{Jencks, Williams, and Coleman}{Jencks
  et~al\mbox{.}}{2009}]%
        {Jencks2009}
\bibfield{author}{\bibinfo{person}{Stephen~F. Jencks}, \bibinfo{person}{Mark~V.
  Williams}, {and} \bibinfo{person}{Eric~A. Coleman}.}
  \bibinfo{year}{2009}\natexlab{}.
\newblock \showarticletitle{{Rehospitalizations among Patients in the Medicare
  Fee-for-Service Program}}.
\newblock \bibinfo{journal}{\emph{New England Journal of Medicine}}
  (\bibinfo{year}{2009}).
\newblock
\showISSN{0028-4793}
\urldef\tempurl%
\url{https://doi.org/10.1056/nejmsa0803563}
\showDOI{\tempurl}


\bibitem[\protect\citeauthoryear{Kemp, Rajkomar, and Dai}{Kemp
  et~al\mbox{.}}{2019}]%
        {Kemp2019}
\bibfield{author}{\bibinfo{person}{Jonas Kemp}, \bibinfo{person}{Alvin
  Rajkomar}, {and} \bibinfo{person}{Andrew~M. Dai}.}
  \bibinfo{year}{2019}\natexlab{}.
\newblock \showarticletitle{{Improved hierarchical patient classification with
  language model pretraining over clinical notes}}.
\newblock \bibinfo{journal}{\emph{arXiv}} (\bibinfo{year}{2019}).
\newblock
\showISSN{23318422}
\showeprint[arxiv]{1909.03039}


\bibitem[\protect\citeauthoryear{Krumholz, Chaudhry, Spertus, Mattera, Hodshon,
  and Herrin}{Krumholz et~al\mbox{.}}{2016}]%
        {Krumholz2016}
\bibfield{author}{\bibinfo{person}{Harlan~M. Krumholz},
  \bibinfo{person}{Sarwat~I. Chaudhry}, \bibinfo{person}{John~A. Spertus},
  \bibinfo{person}{Jennifer~A. Mattera}, \bibinfo{person}{Beth Hodshon}, {and}
  \bibinfo{person}{Jeph Herrin}.} \bibinfo{year}{2016}\natexlab{}.
\newblock \showarticletitle{{Do non-clinical factors improve prediction of
  readmission risk?. Results from the Tele-HF study}}.
\newblock \bibinfo{journal}{\emph{JACC: Heart Failure}} (\bibinfo{year}{2016}).
\newblock
\showISSN{22131779}
\urldef\tempurl%
\url{https://doi.org/10.1016/j.jchf.2015.07.017}
\showDOI{\tempurl}


\bibitem[\protect\citeauthoryear{KS and Sangeetha}{KS and Sangeetha}{2019}]%
        {KS2019}
\bibfield{author}{\bibinfo{person}{Kalyan KS} {and} \bibinfo{person}{S.
  Sangeetha}.} \bibinfo{year}{2019}\natexlab{}.
\newblock \showarticletitle{{SECNLP: A Survey of Embeddings in Clinical Natural
  Language Processing}}.
\newblock \bibinfo{journal}{\emph{arXiv}} (\bibinfo{date}{mar}
  \bibinfo{year}{2019}), \bibinfo{pages}{1--45}.
\newblock
\showISSN{23318422}
\urldef\tempurl%
\url{https://doi.org/10.1016/j.jbi.2019.103323}
\showDOI{\tempurl}
\showeprint[arxiv]{1903.01039}


\bibitem[\protect\citeauthoryear{Lei, Zhou, Zhai, Zhang, Fang, He, and Gao}{Lei
  et~al\mbox{.}}{2019}]%
        {Lei2019}
\bibfield{author}{\bibinfo{person}{Liqi Lei}, \bibinfo{person}{Yangming Zhou},
  \bibinfo{person}{Jie Zhai}, \bibinfo{person}{Le Zhang},
  \bibinfo{person}{Zhijia Fang}, \bibinfo{person}{Ping He}, {and}
  \bibinfo{person}{Ju Gao}.} \bibinfo{year}{2019}\natexlab{}.
\newblock \showarticletitle{{An Effective Patient Representation Learning for
  Time-series Prediction Tasks Based on EHRs}}. In
  \bibinfo{booktitle}{\emph{Proceedings - 2018 IEEE International Conference on
  Bioinformatics and Biomedicine, BIBM 2018}}. \bibinfo{publisher}{IEEE},
  \bibinfo{pages}{885--892}.
\newblock
\showISBNx{9781538654880}
\urldef\tempurl%
\url{https://doi.org/10.1109/BIBM.2018.8621542}
\showDOI{\tempurl}


\bibitem[\protect\citeauthoryear{Li, Rao, Solares, Hassaine, Ramakrishnan,
  Canoy, Zhu, Rahimi, and Salimi-Khorshidi}{Li et~al\mbox{.}}{2020}]%
        {Li2020}
\bibfield{author}{\bibinfo{person}{Yikuan Li}, \bibinfo{person}{Shishir Rao},
  \bibinfo{person}{Jos{\'{e}} Roberto~Ayala Solares},
  \bibinfo{person}{Abdelaali Hassaine}, \bibinfo{person}{Rema Ramakrishnan},
  \bibinfo{person}{Dexter Canoy}, \bibinfo{person}{Yajie Zhu},
  \bibinfo{person}{Kazem Rahimi}, {and} \bibinfo{person}{Gholamreza
  Salimi-Khorshidi}.} \bibinfo{year}{2020}\natexlab{}.
\newblock \showarticletitle{{BEHRT: Transformer for Electronic Health
  Records}}.
\newblock \bibinfo{journal}{\emph{Scientific Reports}} \bibinfo{volume}{10},
  \bibinfo{number}{1} (\bibinfo{year}{2020}), \bibinfo{pages}{1--12}.
\newblock
\showISSN{20452322}
\urldef\tempurl%
\url{https://doi.org/10.1038/s41598-020-62922-y}
\showDOI{\tempurl}


\bibitem[\protect\citeauthoryear{Loshchilov and Hutter}{Loshchilov and
  Hutter}{2017}]%
        {loshchilov2017sgdr}
\bibfield{author}{\bibinfo{person}{Ilya Loshchilov} {and}
  \bibinfo{person}{Frank Hutter}.} \bibinfo{year}{2017}\natexlab{}.
\newblock \bibinfo{title}{SGDR: Stochastic Gradient Descent with Warm
  Restarts}.
\newblock
\newblock
\showeprint[arxiv]{1608.03983}~[cs.LG]


\bibitem[\protect\citeauthoryear{Loshchilov and Hutter}{Loshchilov and
  Hutter}{2019}]%
        {loshchilov2019decoupled}
\bibfield{author}{\bibinfo{person}{Ilya Loshchilov} {and}
  \bibinfo{person}{Frank Hutter}.} \bibinfo{year}{2019}\natexlab{}.
\newblock \bibinfo{title}{Decoupled Weight Decay Regularization}.
\newblock
\newblock
\showeprint[arxiv]{1711.05101}~[cs.LG]


\bibitem[\protect\citeauthoryear{Ma, Chitta, You, Zhou, Xiao, and Gao}{Ma
  et~al\mbox{.}}{2018a}]%
        {Ma2018a}
\bibfield{author}{\bibinfo{person}{Fenglong Ma}, \bibinfo{person}{Radha
  Chitta}, \bibinfo{person}{Quanzeng You}, \bibinfo{person}{Jing Zhou},
  \bibinfo{person}{Houping Xiao}, {and} \bibinfo{person}{Jing Gao}.}
  \bibinfo{year}{2018}\natexlab{a}.
\newblock \showarticletitle{{KAME: Knowledge-based attention model for
  diagnosis prediction in healthcare}}.
\newblock \bibinfo{journal}{\emph{International Conference on Information and
  Knowledge Management, Proceedings}} (\bibinfo{year}{2018}),
  \bibinfo{pages}{743--752}.
\newblock
\showISBNx{9781450360142}
\urldef\tempurl%
\url{https://doi.org/10.1145/3269206.3271701}
\showDOI{\tempurl}


\bibitem[\protect\citeauthoryear{Ma, Chitta, Zhou, You, Sun, and Gao}{Ma
  et~al\mbox{.}}{2017}]%
        {Ma2017}
\bibfield{author}{\bibinfo{person}{Fenglong Ma}, \bibinfo{person}{Radha
  Chitta}, \bibinfo{person}{Jing Zhou}, \bibinfo{person}{Quanzeng You},
  \bibinfo{person}{Tong Sun}, {and} \bibinfo{person}{Jing Gao}.}
  \bibinfo{year}{2017}\natexlab{}.
\newblock \showarticletitle{{Dipole: Diagnosis prediction in healthcare via
  attention-based bidirectional recurrent neural networks}}.
\newblock \bibinfo{journal}{\emph{Proceedings of the ACM SIGKDD International
  Conference on Knowledge Discovery and Data Mining}}  \bibinfo{volume}{Part
  F1296} (\bibinfo{year}{2017}), \bibinfo{pages}{1903--1911}.
\newblock
\showISBNx{9781450348874}
\urldef\tempurl%
\url{https://doi.org/10.1145/3097983.3098088}
\showDOI{\tempurl}
\showeprint[arxiv]{1706.05764}


\bibitem[\protect\citeauthoryear{Ma, Xiao, and Wang}{Ma et~al\mbox{.}}{2018b}]%
        {Ma2018}
\bibfield{author}{\bibinfo{person}{Tengfei Ma}, \bibinfo{person}{Cao Xiao},
  {and} \bibinfo{person}{Fei Wang}.} \bibinfo{year}{2018}\natexlab{b}.
\newblock \showarticletitle{{Health-ATM: A deep architecture for multifaceted
  patient health record representation and risk prediction}}.
\newblock \bibinfo{journal}{\emph{SIAM International Conference on Data Mining,
  SDM 2018}} (\bibinfo{year}{2018}), \bibinfo{pages}{261--269}.
\newblock
\urldef\tempurl%
\url{https://doi.org/10.1137/1.9781611975321.30}
\showDOI{\tempurl}


\bibitem[\protect\citeauthoryear{Min, Yu, and Wang}{Min et~al\mbox{.}}{2019}]%
        {Min2019}
\bibfield{author}{\bibinfo{person}{Xu Min}, \bibinfo{person}{Bin Yu}, {and}
  \bibinfo{person}{Fei Wang}.} \bibinfo{year}{2019}\natexlab{}.
\newblock \showarticletitle{{Predictive Modeling of the Hospital Readmission
  Risk from Patients' Claims Data Using Machine Learning: A Case Study on
  COPD}}.
\newblock \bibinfo{journal}{\emph{Scientific Reports}} (\bibinfo{year}{2019}).
\newblock
\showISSN{20452322}
\urldef\tempurl%
\url{https://doi.org/10.1038/s41598-019-39071-y}
\showDOI{\tempurl}


\bibitem[\protect\citeauthoryear{Padhukasahasram, Reddy, Li, and
  Lanfear}{Padhukasahasram et~al\mbox{.}}{2015}]%
        {Padhukasahasram2015}
\bibfield{author}{\bibinfo{person}{Badri Padhukasahasram},
  \bibinfo{person}{Chandan~K. Reddy}, \bibinfo{person}{Yan Li}, {and}
  \bibinfo{person}{David~E. Lanfear}.} \bibinfo{year}{2015}\natexlab{}.
\newblock \showarticletitle{{Joint impact of clinical and behavioral variables
  on the risk of unplanned readmission and death after a heart failure
  hospitalization}}.
\newblock \bibinfo{journal}{\emph{PLoS ONE}} (\bibinfo{year}{2015}).
\newblock
\showISSN{19326203}
\urldef\tempurl%
\url{https://doi.org/10.1371/journal.pone.0129553}
\showDOI{\tempurl}


\bibitem[\protect\citeauthoryear{Pereira, Choquet, Perozziello, Wargon,
  Juillien, Colosi, Hellmann, Ranaivoson, and Casalino}{Pereira
  et~al\mbox{.}}{2015}]%
        {Pereira2015}
\bibfield{author}{\bibinfo{person}{Laurent Pereira},
  \bibinfo{person}{Christophe Choquet}, \bibinfo{person}{Anne Perozziello},
  \bibinfo{person}{Mathias Wargon}, \bibinfo{person}{Gaelle Juillien},
  \bibinfo{person}{Luisa Colosi}, \bibinfo{person}{Romain Hellmann},
  \bibinfo{person}{Michel Ranaivoson}, {and} \bibinfo{person}{Enrique
  Casalino}.} \bibinfo{year}{2015}\natexlab{}.
\newblock \showarticletitle{{Unscheduled-return-visits after an emergency
  department (ED) attendance and clinical link between both visits in patients
  aged 75 years and over: A prospective observational study}}.
\newblock \bibinfo{journal}{\emph{PLoS ONE}} (\bibinfo{year}{2015}).
\newblock
\showISSN{19326203}
\urldef\tempurl%
\url{https://doi.org/10.1371/journal.pone.0123803}
\showDOI{\tempurl}


\bibitem[\protect\citeauthoryear{Rajkomar, Oren, Chen, Dai, Hajaj, Hardt, Liu,
  Liu, Marcus, Sun, Sundberg, Yee, Zhang, Zhang, Flores, Duggan, Irvine, Le,
  Litsch, Mossin, Tansuwan, Wang, Wexler, Wilson, Ludwig, Volchenboum, Chou,
  Pearson, Madabushi, Shah, Butte, Howell, Cui, Corrado, and Dean}{Rajkomar
  et~al\mbox{.}}{2018}]%
        {Rajkomar2018}
\bibfield{author}{\bibinfo{person}{Alvin Rajkomar}, \bibinfo{person}{Eyal
  Oren}, \bibinfo{person}{Kai Chen}, \bibinfo{person}{Andrew~M. Dai},
  \bibinfo{person}{Nissan Hajaj}, \bibinfo{person}{Michaela Hardt},
  \bibinfo{person}{Peter~J. Liu}, \bibinfo{person}{Xiaobing Liu},
  \bibinfo{person}{Jake Marcus}, \bibinfo{person}{Mimi Sun},
  \bibinfo{person}{Patrik Sundberg}, \bibinfo{person}{Hector Yee},
  \bibinfo{person}{Kun Zhang}, \bibinfo{person}{Yi Zhang},
  \bibinfo{person}{Gerardo Flores}, \bibinfo{person}{Gavin~E. Duggan},
  \bibinfo{person}{Jamie Irvine}, \bibinfo{person}{Quoc Le},
  \bibinfo{person}{Kurt Litsch}, \bibinfo{person}{Alexander Mossin},
  \bibinfo{person}{Justin Tansuwan}, \bibinfo{person}{De Wang},
  \bibinfo{person}{James Wexler}, \bibinfo{person}{Jimbo Wilson},
  \bibinfo{person}{Dana Ludwig}, \bibinfo{person}{Samuel~L. Volchenboum},
  \bibinfo{person}{Katherine Chou}, \bibinfo{person}{Michael Pearson},
  \bibinfo{person}{Srinivasan Madabushi}, \bibinfo{person}{Nigam~H. Shah},
  \bibinfo{person}{Atul~J. Butte}, \bibinfo{person}{Michael~D. Howell},
  \bibinfo{person}{Claire Cui}, \bibinfo{person}{Greg~S. Corrado}, {and}
  \bibinfo{person}{Jeffrey Dean}.} \bibinfo{year}{2018}\natexlab{}.
\newblock \showarticletitle{{Scalable and accurate deep learning with
  electronic health records}}.
\newblock \bibinfo{journal}{\emph{npj Digital Medicine}} \bibinfo{volume}{1},
  \bibinfo{number}{1} (\bibinfo{date}{dec} \bibinfo{year}{2018}),
  \bibinfo{pages}{18}.
\newblock
\showISSN{2398-6352}
\urldef\tempurl%
\url{https://doi.org/10.1038/s41746-018-0029-1}
\showDOI{\tempurl}
\showeprint[arxiv]{1801.07860}


\bibitem[\protect\citeauthoryear{Rasmy, Xiang, Xie, Tao, and Zhi}{Rasmy
  et~al\mbox{.}}{2020}]%
        {Rasmy2020}
\bibfield{author}{\bibinfo{person}{Laila Rasmy}, \bibinfo{person}{Yang Xiang},
  \bibinfo{person}{Ziqian Xie}, \bibinfo{person}{Cui Tao}, {and}
  \bibinfo{person}{Degui Zhi}.} \bibinfo{year}{2020}\natexlab{}.
\newblock \showarticletitle{{Med-BERT: pre-trained contextualized embeddings on
  large-scale structured electronic health records for disease prediction}}.
\newblock \bibinfo{journal}{\emph{arXiv}} (\bibinfo{date}{may}
  \bibinfo{year}{2020}), \bibinfo{pages}{6}.
\newblock
\showeprint[arxiv]{2005.12833}
\urldef\tempurl%
\url{http://arxiv.org/abs/2005.12833}
\showURL{%
\tempurl}


\bibitem[\protect\citeauthoryear{Roy, Teredesai, Zolfaghar, Liu, Hazel, Newman,
  and Marinez}{Roy et~al\mbox{.}}{2015}]%
        {Roy2015}
\bibfield{author}{\bibinfo{person}{Senjuti~Basu Roy}, \bibinfo{person}{Ankur
  Teredesai}, \bibinfo{person}{Kiyana Zolfaghar}, \bibinfo{person}{Rui Liu},
  \bibinfo{person}{David Hazel}, \bibinfo{person}{Stacey Newman}, {and}
  \bibinfo{person}{Albert Marinez}.} \bibinfo{year}{2015}\natexlab{}.
\newblock \showarticletitle{{Dynamic hierarchical classification for patient
  risk-of-readmission}}.
\newblock \bibinfo{journal}{\emph{Proceedings of the ACM SIGKDD International
  Conference on Knowledge Discovery and Data Mining}}
  \bibinfo{volume}{2015-August} (\bibinfo{year}{2015}),
  \bibinfo{pages}{1691--1700}.
\newblock
\showISBNx{9781450336642}
\urldef\tempurl%
\url{https://doi.org/10.1145/2783258.2788585}
\showDOI{\tempurl}


\bibitem[\protect\citeauthoryear{Shang, Ma, Xiao, and Sun}{Shang
  et~al\mbox{.}}{2019}]%
        {Shang2019}
\bibfield{author}{\bibinfo{person}{Junyuan Shang}, \bibinfo{person}{Tengfei
  Ma}, \bibinfo{person}{Cao Xiao}, {and} \bibinfo{person}{Jimeng Sun}.}
  \bibinfo{year}{2019}\natexlab{}.
\newblock \showarticletitle{{Pre-training of Graph Augmented Transformers for
  Medication Recommendation}}.
\newblock \bibinfo{journal}{\emph{IJCAI International Joint Conference on
  Artificial Intelligence}}  \bibinfo{volume}{2019-Augus} (\bibinfo{date}{jun}
  \bibinfo{year}{2019}), \bibinfo{pages}{5953--5959}.
\newblock
\showISBNx{9780999241141}
\showISSN{10450823}
\urldef\tempurl%
\url{https://doi.org/10.24963/ijcai.2019/825}
\showDOI{\tempurl}
\showeprint[arxiv]{1906.00346}


\bibitem[\protect\citeauthoryear{Shickel, Tighe, Bihorac, and Rashidi}{Shickel
  et~al\mbox{.}}{2018}]%
        {Shickel2018}
\bibfield{author}{\bibinfo{person}{Benjamin Shickel},
  \bibinfo{person}{Patrick~James Tighe}, \bibinfo{person}{Azra Bihorac}, {and}
  \bibinfo{person}{Parisa Rashidi}.} \bibinfo{year}{2018}\natexlab{}.
\newblock \showarticletitle{{Deep EHR: A Survey of Recent Advances in Deep
  Learning Techniques for Electronic Health Record (EHR) Analysis}}.
\newblock \bibinfo{journal}{\emph{IEEE Journal of Biomedical and Health
  Informatics}} \bibinfo{volume}{22}, \bibinfo{number}{5}
  (\bibinfo{year}{2018}), \bibinfo{pages}{1589--1604}.
\newblock
\showISSN{21682194}
\urldef\tempurl%
\url{https://doi.org/10.1109/JBHI.2017.2767063}
\showDOI{\tempurl}
\showeprint[arxiv]{1706.03446}


\bibitem[\protect\citeauthoryear{Si, Du, Li, Jiang, Miller, Wang, Zheng, and
  Roberts}{Si et~al\mbox{.}}{2020}]%
        {Si2020}
\bibfield{author}{\bibinfo{person}{Yuqi Si}, \bibinfo{person}{Jingcheng Du},
  \bibinfo{person}{Zhao Li}, \bibinfo{person}{Xiaoqian Jiang},
  \bibinfo{person}{Timothy Miller}, \bibinfo{person}{Fei Wang},
  \bibinfo{person}{W.~Jim Zheng}, {and} \bibinfo{person}{Kirk Roberts}.}
  \bibinfo{year}{2020}\natexlab{}.
\newblock \showarticletitle{{Deep Representation Learning of Patient Data from
  Electronic Health Records (EHR): A Systematic Review}}.
\newblock \bibinfo{journal}{\emph{arXiv}} (\bibinfo{date}{oct}
  \bibinfo{year}{2020}), \bibinfo{pages}{1--47}.
\newblock
\showISSN{23318422}
\urldef\tempurl%
\url{https://doi.org/10.1016/j.jbi.2020.103671}
\showDOI{\tempurl}
\showeprint[arxiv]{2010.02809}


\bibitem[\protect\citeauthoryear{Si and Roberts}{Si and Roberts}{2019}]%
        {Si2019}
\bibfield{author}{\bibinfo{person}{Yuqi Si} {and} \bibinfo{person}{Kirk
  Roberts}.} \bibinfo{year}{2019}\natexlab{}.
\newblock \showarticletitle{{Deep Patient Representation of Clinical Notes via
  Multi-Task Learning for Mortality Prediction.}}
\newblock \bibinfo{journal}{\emph{AMIA Joint Summits on Translational Science
  proceedings. AMIA Joint Summits on Translational Science}}
  \bibinfo{volume}{2019} (\bibinfo{year}{2019}), \bibinfo{pages}{779--788}.
\newblock
\showISSN{2153-4063}
\urldef\tempurl%
\url{http://www.ncbi.nlm.nih.gov/pubmed/31259035{\%}0Ahttp://www.pubmedcentral.nih.gov/articlerender.fcgi?artid=PMC6568068}
\showURL{%
\tempurl}


\bibitem[\protect\citeauthoryear{Steinberg, Jung, Fries, Corbin, Pfohl, and
  Shah}{Steinberg et~al\mbox{.}}{2021}]%
        {Steinberg2021}
\bibfield{author}{\bibinfo{person}{Ethan Steinberg}, \bibinfo{person}{Ken
  Jung}, \bibinfo{person}{Jason~A. Fries}, \bibinfo{person}{Conor~K. Corbin},
  \bibinfo{person}{Stephen~R. Pfohl}, {and} \bibinfo{person}{Nigam~H. Shah}.}
  \bibinfo{year}{2021}\natexlab{}.
\newblock \showarticletitle{{Language models are an effective representation
  learning technique for electronic health record data}}.
\newblock \bibinfo{journal}{\emph{Journal of Biomedical Informatics}}
  \bibinfo{volume}{113} (\bibinfo{year}{2021}), \bibinfo{pages}{1--21}.
\newblock
\showISSN{15320464}
\urldef\tempurl%
\url{https://doi.org/10.1016/j.jbi.2020.103637}
\showDOI{\tempurl}
\showeprint[arxiv]{2001.05295}


\bibitem[\protect\citeauthoryear{Tran, Nguyen, Phung, and Venkatesh}{Tran
  et~al\mbox{.}}{2015}]%
        {Tran2015}
\bibfield{author}{\bibinfo{person}{Truyen Tran}, \bibinfo{person}{Tu~Dinh
  Nguyen}, \bibinfo{person}{Dinh Phung}, {and} \bibinfo{person}{Svetha
  Venkatesh}.} \bibinfo{year}{2015}\natexlab{}.
\newblock \showarticletitle{{Learning vector representation of medical objects
  via EMR-driven nonnegative restricted Boltzmann machines (eNRBM)}}.
\newblock \bibinfo{journal}{\emph{Journal of Biomedical Informatics}}
  \bibinfo{volume}{54} (\bibinfo{year}{2015}), \bibinfo{pages}{96--105}.
\newblock
\showISSN{15320464}
\urldef\tempurl%
\url{https://doi.org/10.1016/j.jbi.2015.01.012}
\showDOI{\tempurl}


\bibitem[\protect\citeauthoryear{Tulloch, David, and Thornicroft}{Tulloch
  et~al\mbox{.}}{2016}]%
        {Tulloch2016}
\bibfield{author}{\bibinfo{person}{A.~D. Tulloch}, \bibinfo{person}{A.~S.
  David}, {and} \bibinfo{person}{G. Thornicroft}.}
  \bibinfo{year}{2016}\natexlab{}.
\newblock \showarticletitle{{Exploring the predictors of early readmission to
  psychiatric hospital}}.
\newblock \bibinfo{journal}{\emph{Epidemiology and Psychiatric Sciences}}
  (\bibinfo{year}{2016}).
\newblock
\showISSN{20457979}
\urldef\tempurl%
\url{https://doi.org/10.1017/S2045796015000128}
\showDOI{\tempurl}


\bibitem[\protect\citeauthoryear{V., S.A., P.J., P., S., M.J., and Y.K.}{V.
  et~al\mbox{.}}{2015}]%
        {V.2015}
\bibfield{author}{\bibinfo{person}{Betihavas V.}, \bibinfo{person}{Frost S.A.},
  \bibinfo{person}{Newton P.J.}, \bibinfo{person}{Macdonald P.},
  \bibinfo{person}{Stewart S.}, \bibinfo{person}{Carrington M.J.}, {and}
  \bibinfo{person}{Chan Y.K.}} \bibinfo{year}{2015}\natexlab{}.
\newblock \bibinfo{title}{{An Absolute Risk Prediction Model to Determine
  Unplanned Cardiovascular Readmissions for Adults with Chronic Heart
  Failure}}.
\newblock
\newblock
\showISBNx{1443-9506}


\bibitem[\protect\citeauthoryear{Vaswani, Shazeer, Parmar, Uszkoreit, Jones,
  Gomez, Kaiser, and Polosukhin}{Vaswani et~al\mbox{.}}{2017}]%
        {Vaswani2017}
\bibfield{author}{\bibinfo{person}{Ashish Vaswani}, \bibinfo{person}{Noam
  Shazeer}, \bibinfo{person}{Niki Parmar}, \bibinfo{person}{Jakob Uszkoreit},
  \bibinfo{person}{Llion Jones}, \bibinfo{person}{Aidan~N. Gomez},
  \bibinfo{person}{{\L}ukasz Kaiser}, {and} \bibinfo{person}{Illia
  Polosukhin}.} \bibinfo{year}{2017}\natexlab{}.
\newblock \showarticletitle{{Attention is all you need}}. In
  \bibinfo{booktitle}{\emph{Advances in Neural Information Processing
  Systems}}.
\newblock
\showISSN{10495258}
\showeprint[arxiv]{1706.03762}


\bibitem[\protect\citeauthoryear{{Wang}, {Cui}, {Chen}, {Avidan}, {Abdallah},
  and {Kronzer}}{{Wang} et~al\mbox{.}}{2018}]%
        {Wang2018}
\bibfield{author}{\bibinfo{person}{H. {Wang}}, \bibinfo{person}{Z. {Cui}},
  \bibinfo{person}{Y. {Chen}}, \bibinfo{person}{M. {Avidan}},
  \bibinfo{person}{A.~B. {Abdallah}}, {and} \bibinfo{person}{A. {Kronzer}}.}
  \bibinfo{year}{2018}\natexlab{}.
\newblock \showarticletitle{Predicting Hospital Readmission via Cost-Sensitive
  Deep Learning}.
\newblock \bibinfo{journal}{\emph{IEEE/ACM Transactions on Computational
  Biology and Bioinformatics}} \bibinfo{volume}{15}, \bibinfo{number}{6}
  (\bibinfo{year}{2018}), \bibinfo{pages}{1968--1978}.
\newblock
\urldef\tempurl%
\url{https://doi.org/10.1109/TCBB.2018.2827029}
\showDOI{\tempurl}


\bibitem[\protect\citeauthoryear{Wang, Porter, Maynard, Bryson, Sun, Lowy,
  McDonell, Frisbee, Nielson, and Fihn}{Wang et~al\mbox{.}}{2012}]%
        {Wang2012}
\bibfield{author}{\bibinfo{person}{Li Wang}, \bibinfo{person}{Brian Porter},
  \bibinfo{person}{Charles Maynard}, \bibinfo{person}{Christopher Bryson},
  \bibinfo{person}{Haili Sun}, \bibinfo{person}{Elliott Lowy},
  \bibinfo{person}{Mary McDonell}, \bibinfo{person}{Kathleen Frisbee},
  \bibinfo{person}{Christopher Nielson}, {and} \bibinfo{person}{Stephan~D.
  Fihn}.} \bibinfo{year}{2012}\natexlab{}.
\newblock \showarticletitle{{Predicting risk of hospitalization or death among
  patients with heart failure in the veterans health administration}}.
\newblock \bibinfo{journal}{\emph{American Journal of Cardiology}}
  (\bibinfo{year}{2012}).
\newblock
\showISSN{00029149}
\urldef\tempurl%
\url{https://doi.org/10.1016/j.amjcard.2012.06.038}
\showDOI{\tempurl}


\bibitem[\protect\citeauthoryear{Xiao, Ma, Dieng, Blei, and Wang}{Xiao
  et~al\mbox{.}}{2018}]%
        {Xiao2018}
\bibfield{author}{\bibinfo{person}{Cao Xiao}, \bibinfo{person}{Tengfei Ma},
  \bibinfo{person}{Adji~B. Dieng}, \bibinfo{person}{David~M. Blei}, {and}
  \bibinfo{person}{Fei Wang}.} \bibinfo{year}{2018}\natexlab{}.
\newblock \showarticletitle{{Readmission prediction via deep contextual
  embedding of clinical concepts}}.
\newblock \bibinfo{journal}{\emph{PLoS ONE}} \bibinfo{volume}{13},
  \bibinfo{number}{4} (\bibinfo{year}{2018}), \bibinfo{pages}{1--15}.
\newblock
\showISBNx{1111111111}
\showISSN{19326203}
\urldef\tempurl%
\url{https://doi.org/10.1371/journal.pone.0195024}
\showDOI{\tempurl}


\bibitem[\protect\citeauthoryear{Xu, Biswal, Deshpande, Maher, and Sun}{Xu
  et~al\mbox{.}}{2018}]%
        {Xu2018}
\bibfield{author}{\bibinfo{person}{Yanbo Xu}, \bibinfo{person}{Siddharth
  Biswal}, \bibinfo{person}{Shriprasad~R. Deshpande}, \bibinfo{person}{Kevin~O.
  Maher}, {and} \bibinfo{person}{Jimeng Sun}.} \bibinfo{year}{2018}\natexlab{}.
\newblock \showarticletitle{{RAIM: Recurrent attentive and intensive model of
  multimodal patient monitoring data}}.
\newblock \bibinfo{journal}{\emph{arXiv}} (\bibinfo{year}{2018}),
  \bibinfo{pages}{2565--2573}.
\newblock
\showISBNx{9781450355520}
\showISSN{23318422}


\bibitem[\protect\citeauthoryear{Yang, Li, Liu, Mei, Xie, Zhao, Xie, and
  Wang}{Yang et~al\mbox{.}}{2017}]%
        {Yang2017}
\bibfield{author}{\bibinfo{person}{Kai Yang}, \bibinfo{person}{Xiang Li},
  \bibinfo{person}{Haifeng Liu}, \bibinfo{person}{Jing Mei},
  \bibinfo{person}{Guotong Xie}, \bibinfo{person}{Junfeng Zhao},
  \bibinfo{person}{Bing Xie}, {and} \bibinfo{person}{Fei Wang}.}
  \bibinfo{year}{2017}\natexlab{}.
\newblock \showarticletitle{{TaGiTeD: Predictive task guided tensor
  decomposition for representation learning from electronic health records}}.
\newblock \bibinfo{journal}{\emph{31st AAAI Conference on Artificial
  Intelligence, AAAI 2017}} (\bibinfo{year}{2017}),
  \bibinfo{pages}{2824--2830}.
\newblock


\bibitem[\protect\citeauthoryear{Yu, Farooq, van Esbroeck, Fung, Anand, and
  Krishnapuram}{Yu et~al\mbox{.}}{2015}]%
        {Yu2015}
\bibfield{author}{\bibinfo{person}{Shipeng Yu}, \bibinfo{person}{Faisal
  Farooq}, \bibinfo{person}{Alexander van Esbroeck}, \bibinfo{person}{Glenn
  Fung}, \bibinfo{person}{Vikram Anand}, {and} \bibinfo{person}{Balaji
  Krishnapuram}.} \bibinfo{year}{2015}\natexlab{}.
\newblock \showarticletitle{{Predicting readmission risk with
  institution-specific prediction models}}.
\newblock \bibinfo{journal}{\emph{Artificial Intelligence in Medicine}}
  (\bibinfo{year}{2015}).
\newblock
\showISSN{18732860}
\urldef\tempurl%
\url{https://doi.org/10.1016/j.artmed.2015.08.005}
\showDOI{\tempurl}


\bibitem[\protect\citeauthoryear{Zhang, Kowsari, Harrison, Lobo, and
  Barnes}{Zhang et~al\mbox{.}}{2018}]%
        {Zhang2018}
\bibfield{author}{\bibinfo{person}{Jinghe Zhang}, \bibinfo{person}{Kamran
  Kowsari}, \bibinfo{person}{James~H. Harrison}, \bibinfo{person}{Jennifer~M.
  Lobo}, {and} \bibinfo{person}{Laura~E. Barnes}.}
  \bibinfo{year}{2018}\natexlab{}.
\newblock \showarticletitle{{Patient2Vec: A Personalized Interpretable Deep
  Representation of the Longitudinal Electronic Health Record}}.
\newblock \bibinfo{journal}{\emph{IEEE Access}}  \bibinfo{volume}{6}
  (\bibinfo{year}{2018}), \bibinfo{pages}{65333--65346}.
\newblock
\showISSN{21693536}
\urldef\tempurl%
\url{https://doi.org/10.1109/ACCESS.2018.2875677}
\showDOI{\tempurl}
\showeprint[arxiv]{1810.04793}


\bibitem[\protect\citeauthoryear{Zhang, Tang, Dodge, Zhou, and Wang}{Zhang
  et~al\mbox{.}}{2019}]%
        {Zhang2019}
\bibfield{author}{\bibinfo{person}{Xi~Sheryl Zhang}, \bibinfo{person}{Fengyi
  Tang}, \bibinfo{person}{Hiroko~H. Dodge}, \bibinfo{person}{Jiayu Zhou}, {and}
  \bibinfo{person}{Fei Wang}.} \bibinfo{year}{2019}\natexlab{}.
\newblock \showarticletitle{{MetaPred: Meta-Learning for Clinical Risk
  Prediction with Limited Patient Electronic Health Records}}. In
  \bibinfo{booktitle}{\emph{Proceedings of the 25th ACM SIGKDD International
  Conference on Knowledge Discovery {\&} Data Mining}}.
  \bibinfo{publisher}{ACM}, \bibinfo{address}{New York, NY, USA},
  \bibinfo{pages}{2487--2495}.
\newblock
\showISBNx{9781450362016}
\showISSN{23318422}
\urldef\tempurl%
\url{https://doi.org/10.1145/3292500.3330779}
\showDOI{\tempurl}


\end{thebibliography}

\end{document}